\documentclass{article} 
\usepackage{iclr2026_conference,times}

\usepackage{amsmath,amsfonts,bm}

\def\eqref#1{equation~\ref{#1}}

\def\1{\bm{1}}

\DeclareMathAlphabet{\mathsfit}{\encodingdefault}{\sfdefault}{m}{sl}
\SetMathAlphabet{\mathsfit}{bold}{\encodingdefault}{\sfdefault}{bx}{n}

\usepackage{hyperref}
\usepackage{url}
\usepackage{booktabs}
\usepackage{multirow}
\usepackage{graphicx}
\usepackage{wrapfig}
\usepackage{subcaption}
\usepackage[capitalize]{cleveref}

\title{\openforgerl{}: Train Harness-native Agents \\ in Any Environment}

\author{Xiao Yu$^1$, \, Baolin Peng$^{3\dagger}$, \, Ruize Xu$^2$, \, Hao Zou$^1$, Qianhui Wu$^3$, \, Hao Cheng$^3$\\ \bf{Wenlin Yao$^3$,} \, \bf{Nikhil Singh$^{2}$,} \, \bf{Zhou Yu$^{1*}$}, \bf{Jianfeng Gao$^{3}$}\thanks{Equal Advisory Contribution; $^\dagger$ Project Lead} \\
$^1$Columbia University \, $^{2}$ Dartmouth College \, $^3$Microsoft Research\\
\texttt{xy2437@columbia.edu,}\,
\texttt{baolinpeng@microsoft.com}
}

\newcommand{\second}[1]{\textcolor{gray}{\textit{#1}}}

\newcommand{\openforge}{\textsc{OpenForge}}
\newcommand{\openforgerl}{\textsc{OpenForge RL}}

\newcommand{\verl}{veRL}
\newcommand{\react}{ReACT}

\newcommand{\openforgeclaw}{OpenForge-Claw}
\newcommand{\openforgegui}{OpenForge-GUI}
\iclrfinalcopy 
\begin{document}

\maketitle
\lhead{Under review as a conference paper at ICLR 2027}

\begin{abstract}
Modern AI agents rely on elaborate \emph{inference harnesses} such as Claude Code, Codex, and OpenClaw to drive multi-turn reasoning, tool use, and access to external systems.
While powerful, these complex harnesses also make agents hard to train end-to-end with open infrastructure, whose SFT/RL stacks cannot natively express stateful, multi-process harness inference.
To address this, we present \textbf{\openforgerl{}}, an open-source framework for training harness-based agents end-to-end in diverse environments.
\openforgerl{} achieves this with a lightweight proxy that serves the harness's model calls while recording them as training data for a standard RL codebase (e.g., \verl{}), and a Kubernetes orchestrator that runs each rollout in its own remote container, together enabling training on \emph{any harness in any environment} at scale.
By decoupling training and inference, \openforgerl{} allows researchers to easily train, study, and improve agents directly in the real harnesses and environments they are deployed with.
We validate our framework across diverse, complex harnesses and environments, spanning tool/claw-based agents and multimodal GUI browser- and computer-use agents.
Using only hundreds to a few thousand tasks, \openforgeclaw{} reaches $31.7$~($\mathrm{pass}^3$) and $55.9$~($\mathrm{pass}@3$) on ClawEval and $33.7$ on QwenClawBench.
\openforgegui{} reaches $37.7$ on OSWorld-Verified, $63.0$ on Online-Mind2Web, and $72.3$ on WebVoyager.
Both outperform open baselines of similar size on nearly all benchmarks, and in the GUI setting match or surpass models several times larger. 
Beyond benchmarks, we analyze how harness choice (e.g., ZeroClaw, OpenClaw, Codex) and RL shape agent behavior.
We find that some harnesses are substantially harder to learn than others, and that RL improves \emph{agentic reliability}, such as self-verification, tool coverage, and completing multi-step plans, though critical abilities such as error recovery remain weak.\footnote{\url{https://aka.ms/OpenForge-RL}}
\end{abstract}

\vspace{-8pt}
\begin{center}
\includegraphics[width=0.88\textwidth]{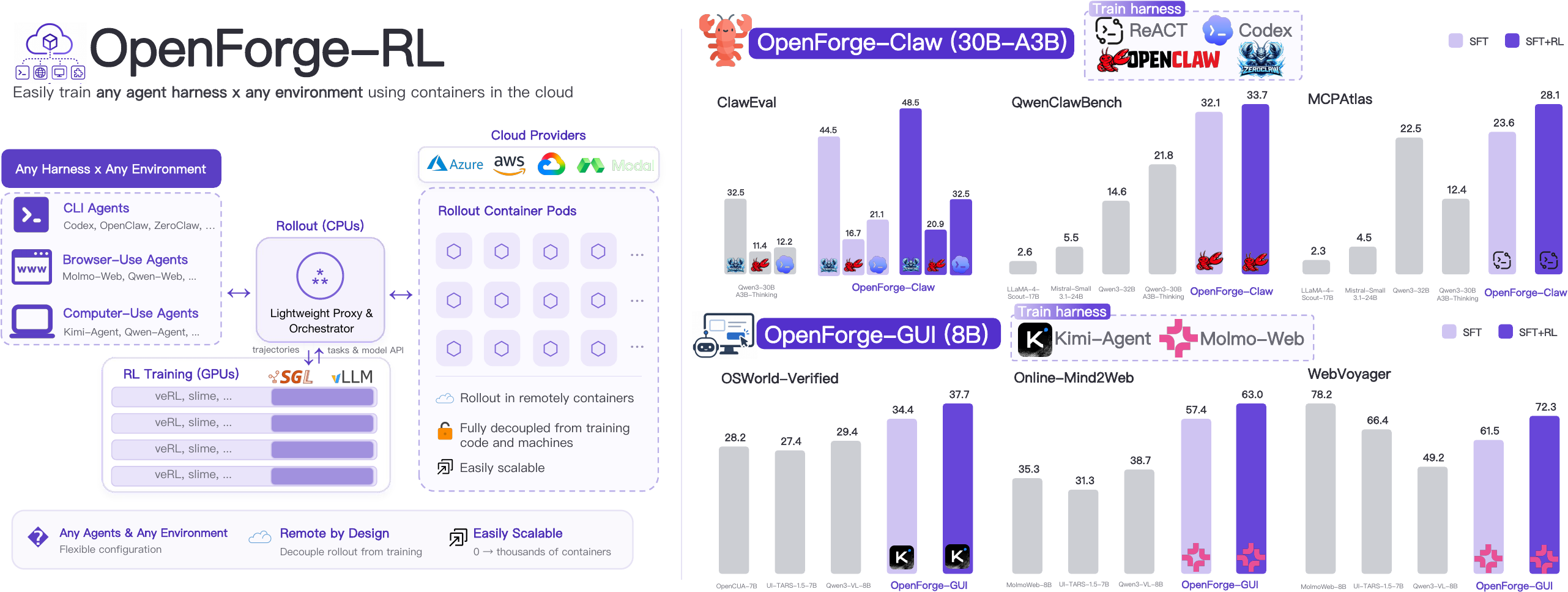}
\captionof{figure}{\emph{Left}: \openforgerl{} builds on Orchard Env \citep{peng2026orchardopensourceagenticmodeling} and connects \emph{any harness $\times$ any environment} to standard RL codebases such as \verl{}, with no train--deploy mismatch. \emph{Right}: \openforge{}-trained models evaluated with six harnesses across six Claw and GUI environments.}
\label{fig:cover}
\end{center}

\section{Introduction}

Modern AI agents are increasingly deployed in complex, open-ended environments, from software engineering \citep{jimenez2024swebenchlanguagemodelsresolve,yang2024sweagent,merrill2026terminalbenchbenchmarkingagentshard} and tool-use \citep{bandi2026mcpatlas,patil2025bfcl} to controlling real web browsers and desktops \citep{zhou2024webarenarealisticwebenvironment,onlinemine2web,xie2024osworldbenchmarkingmultimodalagents}.
To operate effectively in these settings, state-of-the-art agents are rarely a bare language or vision-language model; instead, they are wrapped in increasingly sophisticated \emph{inference harnesses}: orchestration scaffolds such as Claude Code \citep{claudecode}, Codex \citep{codex}, and OpenClaw \citep{openclaw} that manage multi-turn interaction, tool use, and context while linking the model to external systems such as MCP servers \citep{mcp}, browsers, and GUIs.
This harness layer is now central to agentic capability: recent progress on agentic benchmarks comes as much from harnesses as from stronger base models \citep{yang2024sweagent,harnessengineering}.

While harness-equipped agents are remarkably capable, improving them \emph{end-to-end} remains out of reach for much of the open research community for two reasons.
First, a harness turns inference into a stateful, multi-process procedure, with nested tool calls, subagents, and long-horizon context, that open training stacks \citep{verl,slime_github,cao2025skyrl} cannot natively express.
Open efforts often need to reimplement a simplified harness for training \citep{wei2025swerladvancingllmreasoning,wang2026openclawrl}, creating a train--deploy mismatch.
Second, running harnesses at scale requires dedicated, containerized environments that cannot be co-located on training nodes, yet most open RL frameworks assume rollouts run locally inside the trainer.
Together, these gaps leave proprietary harness-based systems increasingly ahead of what the research community can train and study.

To close these gaps, we introduce \textbf{\openforgerl{}}, an open framework that makes it accessible to train harness-based agents end-to-end.
At its core, \openforgerl{} addresses these obstacles with two lightweight components (see \Cref{fig:cover} left or \Cref{fig:rollout-arch}).
First, a lightweight proxy abstracts the harness's inference process and decouples it from training, so that \emph{any} harness can run its own arbitrary inference. Paired with automatic trajectory reconstruction, the recorded prompt-response pairs become standard samples compatible with any RL codebase (e.g., \verl{} \citep{verl}).
Second, a Kubernetes orchestrator, following \citet{peng2026orchardopensourceagenticmodeling}, launches each rollout as a remote container on cloud providers such as Microsoft Azure, scaling elastically to many concurrent environments.
Together, this reuses the rich harness ecosystem directly, spans environments from tool use to GUIs, and remains agnostic to the underlying RL algorithm.

We validate \openforgerl{} across a broad spectrum of complex agentic settings, ranging from daily tool-use and claw-based agents to multimodal GUI browser- and computer-use agents.
In the daily tool-use setting, \textbf{OpenForge-Claw} (30B-A3B MoE) trains on 3 popular harnesses (ZeroClaw, OpenClaw, and Codex) in addition to the standard \react{} loop, and reaches $31.7$~($\mathrm{pass}^3$) and $55.9$~($\mathrm{pass}@3$) on ClawEval \citep{ye2026clawevaltrustworthyevaluationautonomous}, $33.7$ on QwenClawBench \citep{qwenclawbench1.1}, and $28.1$ on MCPAtlas \citep{bandi2026mcpatlas}.
In the GUI setting, \textbf{OpenForge-GUI} (8B) trains on (a modified version of) Kimi-Agent \citep{xie2024osworldbenchmarkingmultimodalagents} and Molmo-Web \citep{gupta2026molmowebopenvisualweb}, and attains $37.7$ on OSWorld-Verified, $63.0$ on Online-Mind2Web, and $72.3$ on WebVoyager, outperforming open baselines of similar scale and matching or surpassing models several times larger.
Crucially, because \openforgerl{} trains agents in their real deployment harnesses, it also lets us study how harness choice and RL training shape agent behavior, a question prior open work could not easily ask (\Cref{sec:Discussion}): some harnesses prove far harder to learn than others, and RL broadly improves \emph{agentic reliability} though abilities such as error recovery remain weak.

In summary, our contributions are threefold:
\begin{itemize}
    \item We introduce \emph{\openforgerl{}}, a scalable training infrastructure that flexibly pairs \emph{any harness with any environment}, connecting real harness ecosystems and remote sandboxes in cloud-service providers to powerful RL codebases (e.g., \verl{})
    \item We conduct a broad empirical study across tool-use/claw and GUI (browser and computer-use) agents, improving over open models of similar size on nearly all benchmarks and, in several cases, over models several times larger.
    \item Enabled by training in real deployment harnesses, we analyze how harness choice and RL shape agents: simpler, better-aligned harnesses are easier to learn; training generalizes across similar harnesses; and RL overall improves \emph{agentic reliability} (e.g., self-verification, tool coverage, and task completion), though error recovery remains challenging to learn.
\end{itemize}


\section{Related Work}
\label{sec:Related Work}

\paragraph{Agents with inference harnesses.}
With the advent of LLM-based coding agents, early work such as SWE-Agent \citep{yang2024sweagent} showed that a carefully designed agent-computer interface substantially improves task success \citep{jimenez2024swebenchlanguagemodelsresolve,xia2024agentlessdemystifyingllmbasedsoftware,wang2025openhandsopenplatformai}. Subsequent harnesses such as Claude Code \citep{claudecode} and Codex \citep{codex} refined this recipe for software engineering, and open-source efforts such as OpenClaw \citep{openclaw} extended it to everyday tasks as well as to GUI tasks \citep{hong2024cogagentvisuallanguagemodel,qin2025uitarspioneeringautomatedgui,xie2024osworldbenchmarkingmultimodalagents,agashe2024agentsopenagentic}.
Rather than developing harnesses, we study how to train LLM- and VLM-based agents end-to-end using these harnesses, aligning agent training with real-world usage so that models can learn in the same setting in which they are deployed.

\paragraph{Training harness-based agents.}
As agents take on increasingly complex environments, training them end-to-end has become a central goal, with reinforcement learning (RL) emerging as the primary tool.
Several open-source RL frameworks, such as \verl{} \citep{verl}, Slime \citep{slime_github}, and more \citep{hu2024openrlhf,cao2025skyrl,fu2026areallargescaleasynchronousreinforcement}, support advanced algorithms (e.g., PPO and GRPO) and asynchronous distributed training.
However, they assume relatively simple rollouts: single-turn generation, or multi-turn interaction with lightweight tool calls such as sandboxed code execution.
These assumptions break for complex inference harnesses, which manage multi-turn interaction, tool use, and context internally, and whose rollouts require containerized environments with dedicated CPU and memory that cannot be co-located on the training nodes at scale.
As a result, initial attempts training agents in these complex environments requires heavily modifing the training loop to fit specific task case-by-case \citep{bai2024digirltraininginthewilddevicecontrol,jin2025searchr1trainingllmsreason,qi2025webrltrainingllmweb,yu2025dynamindlearningsimulateexperience} --- making the codebase hard to extend and maintain.
Our work presents a first step towards designing a training flow that natively supports complex harnesses and environments at scale, while remaining agnostic to the underlying RL framework\footnote{Concurrent work such as Polar~\citep{xu2026polar} proposes a similar approach but focuses on software-engineering tasks (SWE-Bench-Verified).
Our work covers a substantially broader setting, spanning text-based claw tasks and vision-based GUI tasks across six benchmarks, and also further analyzes how harness choice and RL shape agent behavior.}

\section{Methods}
\label{sec:Methods}

\subsection{Notation}
\label{subsec:Notation}
Completing tasks in complex, long-horizon environments is commonly formulated as a Markov Decision Process (MDP) $\left\langle \mathcal{S}, \mathcal{A}, \mathcal{T}, \mathcal{R}, \gamma \right\rangle$.
At each step $t$ of a multi-step task, an LLM-based agent $\pi_\theta$ receives a task instruction together with an observation\footnote{Technically, any input to the agent from our environments is an observation (as in a POMDP); to simplify notation, we use $s$ to denote input data received from the environment.} $s_{t} \sim \mathcal{S}$, generates an action $a_{t} \sim \pi(\cdot | s_t)$, and transitions to the next observation $s_{t+1} \sim \mathcal{S}$.
Under naive inference (e.g., \react{} \citet{yao2023reactsynergizingreasoningacting}), the agent is given the raw interaction history $\left\langle s_{t-H}, a_{t-H}, \ldots, s_t\right\rangle$ and reasons before emitting the next action $a_{t}$.
This loop repeats until the task is completed or a maximum number of steps is reached, at which point a terminal reward $r_{T} \sim \mathcal{R}(a_T,s_T)$ indicates success or failure.
However, for complex tasks such as coding and GUI control, prior work has shown that carefully designed tools and advanced control flows---subagents, skills, and planning modes---substantially improve agent performance \citep{wang2023voyageropenendedembodiedagent,wu2023autogenenablingnextgenllm,agashe2024agentsopenagentic}.
In practice, the agent is therefore wrapped in a harness $\mathcal{H}(\pi)$ that supplies these tools and control flows internally, so that its effective context is \emph{no longer} the raw interaction history.
To abstract away this complexity, we denote (i) each prompt-response pair produced during harness inference as $(\mathcal{H}(s_t), a_t)$, and (ii) the full trajectory as an unordered collection $\tau = \left\langle (\mathcal{H}(s_0), a_0), (\mathcal{H}(s_1), a_1), \ldots, (\mathcal{H}(s_T), a_T) \right\rangle$.


\subsection{\openforgerl}
\label{subsec:openforgerl}
While the emerging agent domains beyond software engineering are increasingly well benchmarked \citep{patil2025bfcl,wu2025mcpmarkbenchmarkstresstestingrealistic,bandi2026mcpatlas}, little open-source infrastructure supports training harness-based agents end-to-end in them.
This is challenging for two reasons.
First, existing RL frameworks assume the trainer has direct access to the model's prompts and full control over generation throughout a rollout; harnesses break this assumption, since their control flows, subagents, and context management are not exposed to the trainer.
Second, harness rollouts need containerized environments with dedicated CPU and memory that cannot be co-located on the training nodes \emph{at scale}.
More broadly, open models are rarely trained in the harnesses they are deployed with, widening the gap to closed-source frontier models.

\begin{figure}[t]
\centering
\includegraphics[width=\textwidth]{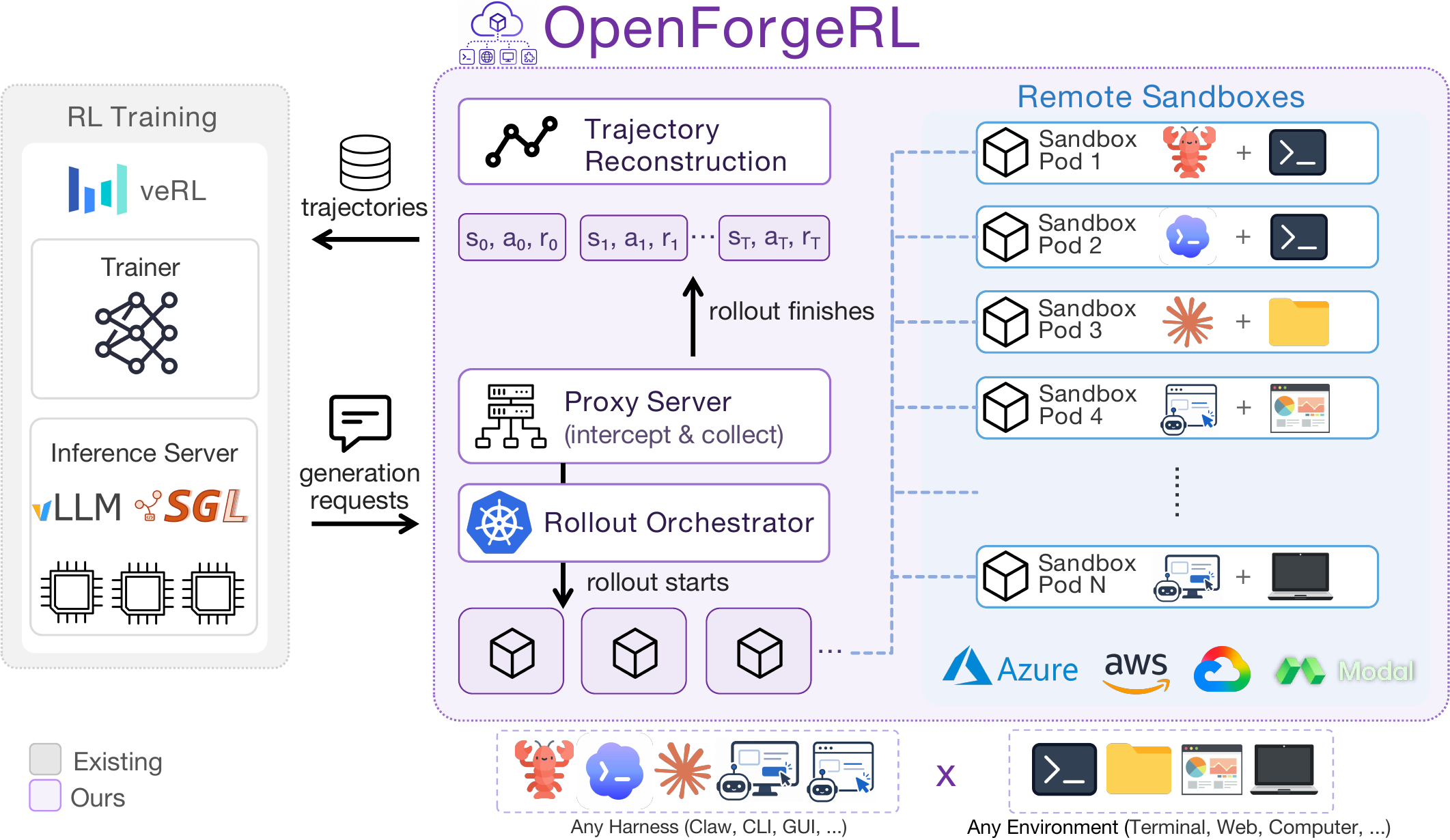}
\caption{Overview of \openforgerl{}. An orchestrator spawns remote sandboxes in which an LLM/VLM interacts with its environment through a harness. A proxy intercepts the harness's LLM calls, routes them to the RL framework's inference engines, and records the exchanged io-pairs as training trajectories.
Supporting a new harness or environment only modifies the sandbox.}
\label{fig:rollout-arch}
\end{figure}

We address both with \openforgerl{}, a plug-and-play rollout interface connecting popular RL frameworks (e.g., \verl{}) to distributed harness rollouts (\Cref{fig:rollout-arch}). Given an inference server (e.g., vLLM \citet{vllm}), \openforgerl{} launches (1) a Kubernetes orchestrator that creates, manages, and deletes rollout container pods on cloud providers such as Microsoft Azure, and (2) a proxy that wraps the inference server and intercepts all generation requests issued by those containers.
When a rollout finishes, the proxy collects the terminal reward $r_T$ (typically task success) and the prompt-response pairs $(s^{\mathcal{H}},a)$ from the container, and reconstructs training samples as:
\begin{equation}
    \label{eq:traj_recon}
    \tau = (s^{\mathcal{H}}_0, a_0, r_0), (s^{\mathcal{H}}_1, a_1, r_1), \ldots, (s^{\mathcal{H}}_T, a_T, r_T);\quad r_{t} = \gamma^{T-t} \cdot r_T
\end{equation}
typically with $\gamma=1.0$. When optimizing with group-based algorithms such as GRPO, we follow \citet{feng2025group} and compute advantage by comparing the average sample rewards in \emph{different} trajectories in the same group.
While offloading rollouts to remote containers enables scaling to many concurrent environments, it raises three practical challenges, which we address below.

\paragraph{Rollout orchestration.}
To manage the lifecycle of many rollout containers, we build on Orchard \citep{peng2026orchardopensourceagenticmodeling} and implement a Kubernetes orchestrator that creates, deletes, and allocates resources for rollout containers on cloud providers such as Microsoft Azure.
This allows us to easily manage and scale the number of concurrent rollouts elastically, without overloading training nodes.
\paragraph{Asynchronous rollout and timeouts.}
Because each rollout runs remotely and outside the trainer's control, a single unresponsive rollout will block the collection of an entire training batch and subsequently stall training.
Capping the number of turns per rollout \citep{jin2025searchr1trainingllmsreason,yu2025dynamindlearningsimulateexperience}, a common safeguard, is unreliable in this setting: some harnesses (e.g., Codex) do not expose a turn limit, and a ``turn'' is defined inconsistently across harnesses.
We instead impose a (generous) wall-clock timeout on each rollout job. When a job exceeds its timeout, we terminate it and return an error signal to the proxy and the trajectory-reconstruction module, so training continues collecting from the remaining rollouts rather than blocking by the terminated one.
\paragraph{Error handling.} Rollouts in these complex environments can fail for reasons unrelated to the policy model, such as network issues, harness crashes, or timeouts. Following DAPO \citep{yu2025dapoopensourcellmreinforcement}, we consider a simple strategy to discard all samples from a trajectory that ended in such an error, since a partial rollout can inject misleading training signal (e.g., a correct prefix that receives a negative reward). Designing better credit assignment for these partial rollouts is a promising direction in this setting, which we leave to future work.

\subsection{Data Synthesis}
\label{subsec:openforgeenv}

\begin{wrapfigure}[21]{r}{0.40\textwidth}
\centering
\vspace{-5em}
\includegraphics[width=\linewidth]{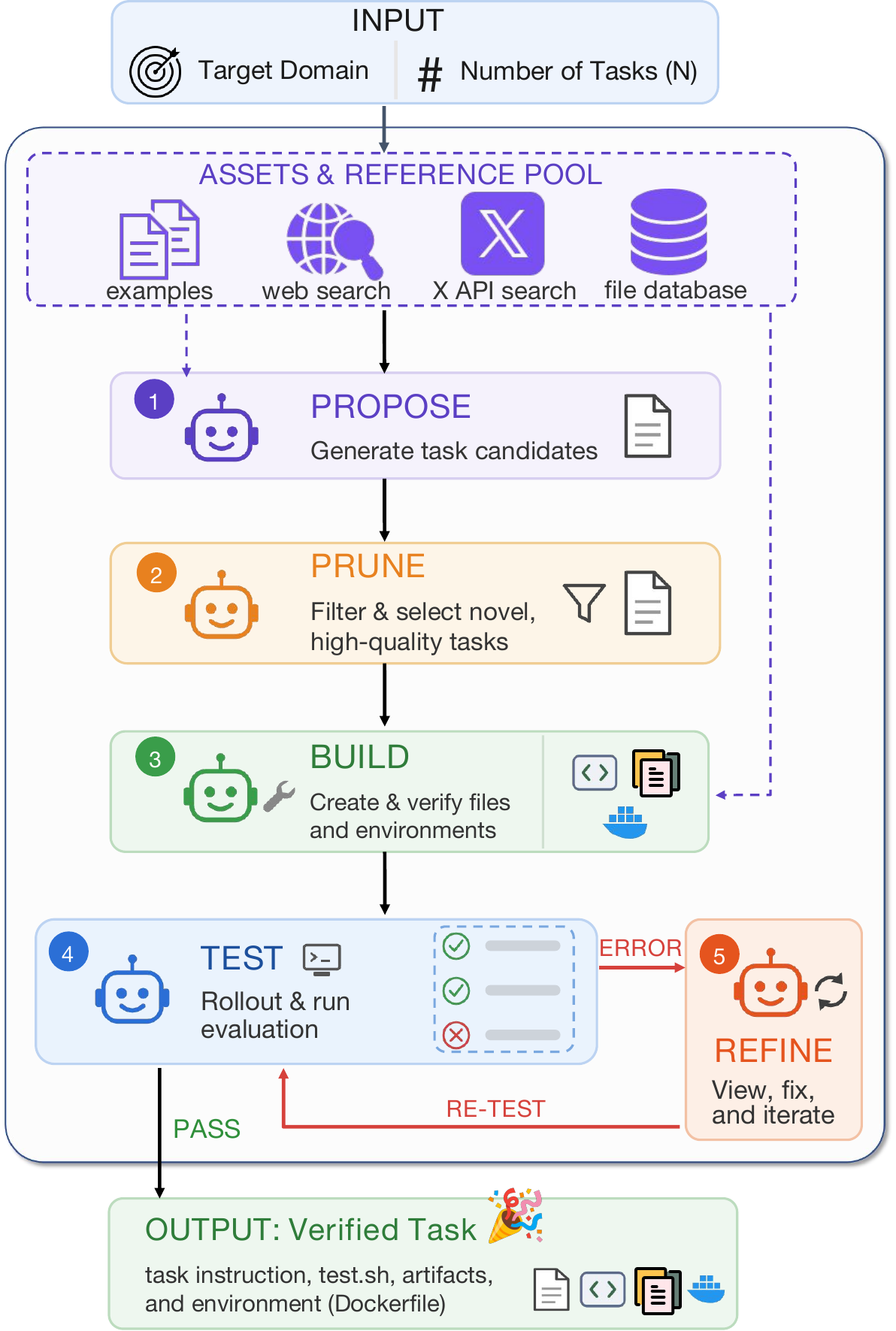}
\caption{Overview of our data/task synthesis pipeline.}
\label{fig:task-synth}
\end{wrapfigure}

This work aims to train harness-based agents end-to-end across diverse environments beyond coding, such as browser- and computer-use.
Unlike coding \citep{badertdinov2026swerebenchv2languageagnosticswe}, these domains offer far fewer training tasks, harnesses, and RL-ready environments.
To help experiment our framework in diverse environments, we therefore also built a simple pipeline to synthesize SFT and RL tasks for such data-scarce domains (e.g., claw/daily tool use and computer use).
We give a high-level overview below and defer full details to \Cref{subsec:Data Synthesis Pipeline}.

The pipeline is built to mimic how a human would curate a task \citep{xie2026agentsynthscalabletaskgeneration,PAE}.
Given a target domain and a number of tasks, it spawns agents in parallel that (1)~\textbf{propose} candidate instructions grounded in realistic scenarios, drawn from the web/X API or a pool of reference assets and instructions; (2)~\textbf{prune} low-quality and duplicate tasks; (3)~\textbf{build} an executable environment and a verifier script for each task; (4)~\textbf{test} the task by rolling out a separate open LLM/VLM; and (5)~\textbf{refine} it by patching defects until it passes all checks.
The test and refine stages are essential: they validate each task end-to-end before it enters the dataset.
\Cref{fig:task-synth} illustrates the pipeline, and \Cref{subsec:Data Synthesis Pipeline} provides full details.
Because each environment is defined by a custom Dockerfile, the pipeline extends naturally from Linux/CLI tasks (e.g., claw) to GUI and computer-use tasks (e.g., rendering virtual displays with Xvfb), and can pre-install arbitrary harnesses such as OpenClaw and Codex.
The resulting tasks and environments support both SFT (via distillation from a stronger model's rollouts) and RL. We will release these data and this pipeline for future use.

\begin{figure}[t]
\centering
\includegraphics[width=\textwidth]{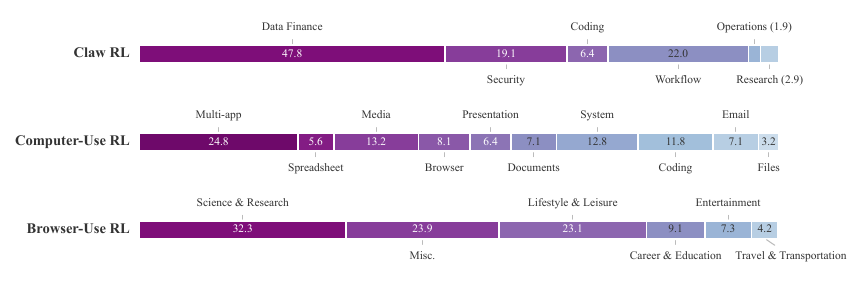}
\caption{Distribution (\%) of tasks used in the Claw, Computer-Use, and Browser-Use domains (top to bottom) for \openforgerl{} training. SFT-task distributions are similar and shown in \Cref{fig:dset-distribution-full}.}
\label{fig:dset-distribution}
\vspace{-2.5em}
\end{figure}

\begin{table}[t]
\centering
\caption{Statistics of SFT and RL data used for Claw and GUI agent training. For GUI, we used a slightly modified version of the Kimi-Agent and Molmo-Web harness (see \Cref{subsec:computeruse evaluation details,subsec:browseruse evaluation details}).}
\label{tab:dset-stats}
\small
\setlength{\tabcolsep}{4pt}
\begin{tabular}{@{}lccc@{}}
\toprule
 & \textbf{Claw} & \textbf{GUI(Computer)} & \textbf{GUI(Browser)} \\
\midrule
Harnesses        & \react{}*,ZeroClaw,OpenClaw,Codex & Kimi-Agent* & MolmoWeb* \\
\# SFT Trajectories     & 892 & 795 & 1,496 \\
\# RL tasks      & 343 & 252 & 900 \\
\bottomrule
\end{tabular}
\end{table}

\section{Experiments}
\label{sec:Experiments}

In this section, we use \openforgerl{} to train agents across a range of harnesses and environments.
We then evaluate the trained agents across six popular benchmarks, spanning both text-based tool-use (claw) and multimodal GUI (browser and computer use) domains.
In the next section, we then study \emph{how} \openforgerl{} training reshapes agent behavior, examining tool-use patterns, generalization to unseen harnesses, and the behavioral changes that RL introduces on top of SFT.


\subsection{Claw Agents}
\label{subsec:Claw Agents}

We first use \openforgerl{} to train agents across \emph{diverse harnesses}.
Here we hold the environment type fixed to text-based tasks, and train and evaluate LLM-based agents on everyday tasks that require diverse tool use, such as reading email, searching a knowledge base, or updating a helpdesk ticket, inside harnesses such as ZeroClaw, OpenClaw, and Codex.

\paragraph{SFT and RL Data} We use our pipeline in \Cref{subsec:openforgeenv} to automatically generate a large pool of tasks and executable environments for both SFT and RL. To seed the proposal stage, we give the pipeline an assets and reference pool drawn from (1) ClawHub access\footnote{The agent can use command lines to lookup popular skills/use cases from \url{https://clawhub.ai/}} and (2) tasks from ZClawBench \citep{zclawbench}, which together supply examples that reflect how people daily usage of claw-based agents.
\Cref{tab:dset-stats} and \Cref{fig:dset-distribution} report the statistics and category distribution of the resulting tasks.
To study the effect of training on diverse harnesses, we pair each task with one of four harnesses for its rollouts: \react{} (solving the task via a simple loop), ZeroClaw, OpenClaw, and Codex.
Training dataset statistics is shown in \Cref{tab:dset-stats} and \Cref{fig:dset-distribution}.
For more details, please see \Cref{subsec:More Details on Claw Data}.

\paragraph{Training Details}
We use Qwen3-30B-A3B-Thinking \citep{yang2025qwen3technicalreport} as the backbone model for training.
For SFT, we distill trajectories from a stronger teacher model (MiniMax-M2.5, \citet{minimaxm25}): we sample $N=3$ rollouts per task and keep only the successful ones for training.
For RL, we continue from the SFT checkpoint and train with GRPO \citep{Guo_2025,shao2024deepseekmathpushinglimitsmathematical}.
To instantiate \openforgerl{}, we use \verl{} \citep{verl} as the training backend and Microsoft Azure as the cloud provider for rollout containers.
We use a batch size of $8$ and a group size of $8$, and train on 8$\times$B200 GPUs.
For additional details, including training hyperparameters and training curves, see \Cref{subsec:Claw Training Details}.

\paragraph{Evaluation Benchmarks}
We evaluate on two popular claw-based benchmarks, ClawEval \citep{ye2026clawevaltrustworthyevaluationautonomous} and QwenClawBench \citep{qwenclawbench1.1}, and one broader tool-call benchmark, MCPAtlas~\citep{bandi2026mcpatlas}.
For ClawEval, we use the 2026-04-08 release and follow the official protocol, reporting $pass^{3}$ and $pass@3$ with the benchmark's default \react{} loop as the harness.
For QwenClawBench, we follow the official implementation and solve the tasks with OpenClaw \citep{openclaw}.
We report $pass@1$ as the main metric.
For MCPAtlas, we use the benchmark's default 20-server configuration, which excludes optional servers requiring third-party credentials or service-specific data initialization for more reproducible results. In total, this includes 89 tasks whose ground-truth expected tool calls are fully supported by this server set.
We then follow the official setting and use the benchmark's LLM-as-judge claim-coverage evaluator, counting a task as successful when its coverage score is at least 0.75.
We report $pass@1$ as the main metric.
For more details on evaluation, please see \Cref{sec:Evaluation Details Appendix}.

\begin{table*}[t]
\centering
\small
\setlength{\tabcolsep}{6pt} 
\begin{tabular}{lcccc}
\toprule
\multirow{2}{*}{\textbf{System}} & \multicolumn{2}{c}{\textbf{ClawEval}} & \textbf{QwenClawBench} & \textbf{MCPAtlas$^{\spadesuit}$}\\
\cmidrule(lr){2-3}\cmidrule(lr){4-4}\cmidrule(lr){5-5}
 & $pass^3$ & $pass@3$ & $pass@1$ & $pass@1$\\
\midrule
\multicolumn{5}{l}{\textit{SOTA Large Language Models}} \\
\midrule
Claude Opus 4.6$^{*\dagger}$   & 70.8 & 80.8 & 59.5 & 76.4\\
GPT 5.4$^{*\dagger}$           & 60.2 & 75.8 & 56.7 & 68.5\\
Gemini 3.1 Pro$^{*}$              & 55.9 & 80.8 & -- & --\\
Qwen3.5 397A17B$^{*}$             & 57.8 & 70.8 & 54.2 & 66.3\\
GLM 5 Turbo$^{*}$      & 52.8 & 73.3 & -- & 67.4\\
MiniMax M2.7$^{*}$      & 49.7 & 72.0 & 50.5 & 50.6\\
MiniMax M2.5                 & 47.2 & 65.2 & 51.2 & 57.3\\
Kimi K2.5$^{*\dagger}$                    & 36.6 & 67.1 & 51.9 & 52.8\\
\midrule
\multicolumn{5}{@{}l}{\textit{Similar-size baselines and our model (30B-A3B; \textasciitilde 3B active)}} \\
\midrule
LLaMA-4-Scout-17B-16E-Instruct & 0.6 & 16.8 & 2.6 & 2.3\\
Mistral-Small-3.1-24B-Instruct & 3.1 & 19.3 & 5.5 & 4.5\\
Qwen3-32B & 6.8 & 31.7 & 14.6 & 22.5\\
Qwen3-30B-A3B-Thinking        & 14.3 & 39.8 & 21.8 & 12.4\\
Qwen3-Coder-30B-A3B-Instruct& 30.4 & 49.7 & 24.3 & 19.1\\
\midrule
\textbf{OpenForge-Claw(SFT)}   & 21.7 & 52.1 & 32.1 & 23.6\\
\textbf{OpenForge-Claw(SFT+RL)} & \textbf{31.7} & \textbf{55.9} & \textbf{33.7} & \textbf{28.1}\\
\bottomrule
\end{tabular}
\caption{Claw-agent performance on Claw-Eval, QwenClawBench, and MCPAtlas. We use the general domain (0408) for Claw-Eval. $^{*}$marks numbers from the Claw-Eval leaderboard. $^{\dagger}$marks numbers from the QwenClawBench leaderboard. 
MCPAtlas$^{\spadesuit}$ results are pass@1 over the 89-task without credential-server configuration.}
\label{tab:claw-results}
\end{table*}

\paragraph{Results}

We present our results in \Cref{tab:claw-results}. Compared to models of similar size (around 30B, or MoE models with $\sim$3B active parameters) and to the untrained Qwen3-30B-A3B-Thinking backbone, our \openforgeclaw{} models achieve superior results across all three benchmarks.
This indicates the effectiveness of our curated tasks and environments, which provide useful learning signal for the model to solve everyday tasks through harnesses such as OpenClaw.
Next, compared to \openforgeclaw{}(SFT), our \openforgeclaw{}(SFT+RL) shows a significant improvement in both robustness ($pass^{3}$ on ClawEval) and average success rate ($pass@1$ on QwenClawBench and MCPAtlas).
This indicates the effectiveness of our \openforgerl{} training infrastructure, which allows the model to explore and learn from online interaction with the environments and harnesses.
For evaluation across different harnesses and generalization to unseen harnesses, please refer to \Cref{subsec:Cross Harness Comparison} and \Cref{subsec:Unseen Harness Generalization}, respectively.

\subsection{GUI agents}
\label{subsec:GUI agents}

Beyond diverse harnesses, we also explore \openforgerl{} training in \emph{diverse environments}, such as multimodal GUI environments that require visual perception and low-level mouse and keyboard control in computer-use and browser-use tasks.

\begin{table*}[t]
\centering
\small
\setlength{\tabcolsep}{4pt} 
\begin{tabular}{lcccc}
\toprule
\textbf{System} & \textbf{\#Steps} & \textbf{OSWorld-Verified} & \textbf{OnlineMind2Web} & \textbf{WebVoyager}\\
\midrule
\multicolumn{5}{l}{\textit{SOTA Vision Language Models}} \\
\midrule
GPT 5.4$^{*\dagger}$           & 100 & 75.0 & 92.8 & --\\
Claude Opus 4.6$^{*\dagger}$   & 100 & 72.7 & 84.0 & --\\
Gemini 3.1 Pro$^{*}$      & 100 &76.2 & -- & --\\
Kimi K2.5$^{*}$    & 100 & 63.3 & 60.4 & 74.3\\
Qwen3-VL 235BA22B$^{*}$   & -- & 38.1 & 63.7 & 66.4\\
OpenCUA-32B$^{*}$  & 50 & 34.1 & -- & --\\
\midrule
\multicolumn{5}{@{}l}{\textit{Similar-size baselines and our model (\textasciitilde 8B)}} \\
\midrule
MolmoWeb-8B$^{\dagger}$  & 30 & -- & 35.3 & \textbf{78.2}\\
OpenCUA-7B$^{*}$   & 50 & 28.2 & -- & --\\
UI-TARS-1.5-7B$^{*\dagger}$   & 100 & 27.4 & 31.3 & 66.4\\
Qwen3-VL-8B & 50 & 29.4 & 38.7 &49.2 \\
\midrule
\textbf{OpenForge-GUI(SFT)}      & 30 &\second{34.4} & \second{57.4} & 61.5\\
\textbf{OpenForge-GUI(SFT+RL)}      & 30 & \textbf{37.7} & \textbf{63.0} & \second{72.3}\\
\bottomrule
\end{tabular}
\caption{GUI-agent performance on OSWorld-verified, OnlineMind2Web, and WebVoyager. All results are $pass@1$. $^{*}$marks numbers reported by model's official technical report; $^{\dagger}$marks numbers reported in \citet{yang2026openwebrldemystifyingonlinemultiturn,gupta2026molmowebopenvisualweb}. Best in shown in \textbf{bold}, second shown in \second{gray}.}
\label{tab:gui-results}
\end{table*}

\paragraph{SFT and RL Data} Following the same procedure as in \Cref{subsec:Claw Agents}, we use our pipeline in \Cref{subsec:openforgeenv} to automatically generate a large pool of tasks and environments.
To seed the proposal stage for computer-use tasks, we give the pipeline a reference and artifact pool consisting of (1) X (formerly Twitter) search API access\footnote{The agent can use the X search API to find real-world use cases of computer-use agents.}, (2) 22k instructions from AgentNet \citep{wang2025opencuaopenfoundationscomputeruse}, and (3) synthetic files and data from Synthetic-Computer-Use \citep{ge2026syntheticcomputersscalelonghorizon} that populate each environment with realistic assets.
To build containerized computer-use environments, we render a virtual display with Xvfb and pre-install a GUI harness (e.g., Kimi-Agent), so that a VLM can control the machine through simulated mouse clicks (e.g., \texttt{left\_click(x,y)}) and keyboard actions (e.g., \texttt{type("text")}).
For browser-use, since synthesizing realistic websites and their databases is impractical, we instead follow OpenWebRL \citep{yang2026openwebrldemystifyingonlinemultiturn} and draw real-website tasks from existing datasets.
Specifically, we (1) start with tasks from WebGym \citep{bai2026webgymscalingtrainingenvironments}; (2) filter tasks that overlap the evaluation benchmarks and also restricting to popular websites; (3) performed deduplication.
This results in an SFT pool of 2500 tasks and 900 RL tasks.
As in OpenWebRL, we prompt GPT-4.1 \citep{gpt-41} as the evaluator to determine task success during both SFT data construction and RL training.
To build containerized browser-use environments, we pre-install a GUI-browser harness (i.e., Molmo-Web) and used remote browser service from Browser-Use \citep{browseruse} to interact with real websites.
Training dataset statistics is shown in \Cref{tab:dset-stats} and \Cref{fig:dset-distribution}.
For more details on our GUI training data, please see \Cref{subsec:More Details on GUI Data}.

\paragraph{Training Details}
We use Qwen3-VL-8B-Thinking \citep{yang2025qwen3technicalreport} as the backbone model for all training.
For SFT, we distill trajectories from a stronger teacher model (Kimi-K2.5, \citet{kimiteam2026kimik25visualagentic}): we sample $N=3$ rollouts per task and keep only the successful ones for training.
For RL, we also follow the previous section - we instantiate \openforgerl{} with \verl{} as the training backend and Microsoft Azure as the cloud provider for rollout containers.
We use GRPO with a batch size of $8$ and a group size of $8$, and train on 8$\times$B200 GPUs.
All models use screenshots as their visual input for both training and evaluation.
For additional details, such as training hyperparameters and training curves, please see \Cref{subsec:GUI Training Details}.

\paragraph{Evaluation Benchmarks}
We evaluate on three popular GUI benchmarks: OSWorld-Verified \citep{xie2024osworldbenchmarkingmultimodalagents} for computer-use environments, and Online-Mind2Web \citep{onlinemine2web} and WebVoyager \citep{he2024webvoyagerbuildingendtoendweb} for browser-use.
For OSWorld-Verified, we follow the official protocol and report the average success rate.
For browser-use, we follow \citet{yang2026openwebrldemystifyingonlinemultiturn,fara7b2025} and report average success rate using the AgentTrek protocal \citep{xu2025agenttrekagenttrajectorysynthesis} with o4-mini \citep{o4-mini} for Online-Mind2Web and the official protocal with GPT-4o \citep{gpt-4o} for WebVoyager.
More details about each benchmark can be found in \Cref{subsec:computeruse evaluation details,subsec:browseruse evaluation details}.

\paragraph{Results}
We present our results in \Cref{tab:gui-results}. Compared to similar-size (around 8B) models that were specifically fine-tuned for computer-use or browser-use (e.g., OpenCUA, UI-TARS, and MolmoWeb), our \openforgegui{} models achieve superior results on nearly all benchmarks.
Notably, while MolmoWeb is trained on over 200k tasks, \openforgegui{} uses only 2.5k tasks yet outperforms it on Online-Mind2Web and stays competitive on WebVoyager.
More importantly, \openforgegui{}(SFT+RL) improves substantially over \openforgegui{}(SFT) on all three benchmarks.
This is a considerably harder test of \openforgerl{} than the text-only setting: every GUI rollout runs a full VLM harness inside a containerized virtual display, perceives the screen visually, and issues long sequences of low-level mouse and keyboard actions against real applications and websites.
These consistent gains show that \openforgerl{} generalizes beyond diverse harnesses to diverse, highly complex environments, training agents end-to-end even under demanding multi-turn, multimodal GUI tasks.


\section{Discussion}
\label{sec:Discussion}

In this section, we analyze how the choice of harness affects learning and what our models learn from harness-based SFT and RL training.
We organize the analysis around three questions.
(1) \emph{Are some harnesses harder to master than others?}~(\Cref{subsec:Cross Harness Comparison}).
(2) \emph{Does training on one harness transfer to unseen harnesses, and does training on several at once help further?}~(\Cref{subsec:Unseen Harness Generalization}).
(3) \emph{What capabilities does RL add on top of SFT?}~(\Cref{subsec:Capability Learned by RL}).
For simplicity, we focus on our study with \openforgeclaw{} models evaluated on the ClawEval benchmark.

\subsection{Evaluation across Different Harnesses}
\label{subsec:Cross Harness Comparison}

\begin{table}[t]
\centering
\caption{Comparing \openforgeclaw{} on ClawEval with different harnesses.}
\label{tab:cross-harness}
\small
\setlength{\tabcolsep}{4pt}
\resizebox{\textwidth}{!}{%
\begin{tabular}{@{}lcccccccc@{}}
\toprule
 & \multicolumn{4}{c}{\textbf{ClawEval($pass@1$)}} & \multicolumn{4}{c}{\textbf{ClawEval($pass@3$)}} \\
\cmidrule(lr){2-5}\cmidrule(lr){6-9}
\textbf{Model} & \react{}* & ZeroClaw & OpenClaw & Codex & \react{}* & ZeroClaw & OpenClaw & Codex \\
\midrule
Qwen3-30B-A3B-Thinking
  & 26.1 & 32.5 & 11.4 & 12.2
  & 39.8 & 44.7 & 19.3 & 18.6 \\


\textbf{Orchard-Claw(SFT)}
  & 36.2 & 44.5 & 16.7 & 21.1
  & 52.1 & 66.5 & 24.2 & 35.4 \\

\textbf{Orchard-Claw(SFT+RL)}
    & \textbf{45.1} & \textbf{48.5} & \textbf{20.9} & \textbf{32.5}
    & \textbf{55.9} & \textbf{67.1} & \textbf{27.8} & \textbf{51.5} \\
\bottomrule
\end{tabular}%
}
\end{table}

First, we study whether some harnesses are harder to learn than others.
In principle, any task is solvable by any harness; in practice, prior work \citep{yang2024sweagent,wang2024executablecodeactionselicit} also find tools and control flows are better aligned with the model's capabilities can often reach higher performance more easily.

We therefore evaluate our trained \openforgeclaw{} models on ClawEval using four harnesses of increasing sophistication: \react{}*, ZeroClaw, OpenClaw, and Codex.
\react{}* is the benchmark's original harness, a \react{}-like loop that repeatedly prompts the model to reason and call tools, and ZeroClaw \citep{zeroclaw} is a lightweight version of OpenClaw that adds an easy way to register new tools.
OpenClaw \citep{openclaw} and Codex \citep{codex} are far more advanced, offering a rich set of built-in tools and control flows, but neither easily supports custom tools; we therefore expose each ClawEval-specific tool to them through \texttt{SKILL.md} files (see \Cref{app:claweval details appendix}).

\Cref{tab:cross-harness} reports the results, from which two trends stand out.
First, the harnesses that support adding custom tools directly (\react{} and ZeroClaw) reach the highest performance.
Second, SFT+RL bring large gains on every harness except OpenClaw, which only has moderate gains while consuming far longer prompts and contexts than the others.
This corroborates prior findings that simpler, better-engineered tools and control flows are essential for agentic performance, and further shows that training on such harnesses advances the model's ability to solve tasks in complex environments.

\subsection{Generalization to Unseen Harness}
\label{subsec:Unseen Harness Generalization}

\begin{wraptable}[10]{r}{0.65\textwidth}
\centering
\vspace{-1.5em}
\caption{
Unseen harness evaluation. All training are from Qwen3-30B-A3B-Thinking.
Deltas are over the base model.
}
\label{tab:unseen-harness}
\small
\setlength{\tabcolsep}{4pt}
\begin{tabular}{@{}lccc@{}}
\toprule
 & \multicolumn{3}{c}{\textbf{ClawEval($pass@1$)}} \\
\cmidrule(lr){2-4}
\textbf{Training (SFT+RL)} & ZeroClaw & OpenClaw & Codex \\
\midrule
None (base)
  & 32.5 & 11.4 & 12.2 \\

ZeroClaw
  & 46.0 {\tiny(+13.5)} & 14.7 {\tiny(+3.3)} & 16.8 {\tiny(+4.6)} \\

\textbf{ZeroClaw+OpenClaw+Codex}
  & \textbf{48.5} {\tiny(+16.0)} & \textbf{20.9} {\tiny(+9.5)} & \textbf{32.5} {\tiny(+20.3)} \\
\bottomrule
\end{tabular}
\end{wraptable}

Next, we ask whether training on one harness transfers to unseen harnesses in evaluation, and whether training on several harnesses at once improves over training on a single one.
We compare two \openforgeclaw{} models built from the same tasks and the same training recipe, differing only in which harnesses generate their rollouts (for both SFT distillation and RL): one trained on ZeroClaw alone, and one trained on ZeroClaw, OpenClaw, and Codex together.
We evaluate both on all three harnesses (\Cref{tab:unseen-harness}). For the ZeroClaw-only model, OpenClaw and Codex are unseen.

We observe two effects.
First, training on a single harness already generalizes to the others: the ZeroClaw-only model improves over the untrained base on the unseen OpenClaw (+3.3) and Codex (+4.6).
Second, training on all three harnesses is best across the board, with the largest gains over the base on the more complex harnesses (+9.5 on OpenClaw and +20.3 on Codex), and it even lifts ZeroClaw itself beyond ZeroClaw-only training (48.5 vs.\ 46.0).
We attribute this to the greater diversity of tool calls and control flows the model sees when trained on multiple harnesses, which makes it more robust across different tools and scenarios.

\subsection{Capability Learned by RL}
\label{subsec:Capability Learned by RL}

\begin{figure}[t]
\centering
\begin{subfigure}[b]{0.56\textwidth}
\centering
\includegraphics[width=\linewidth]{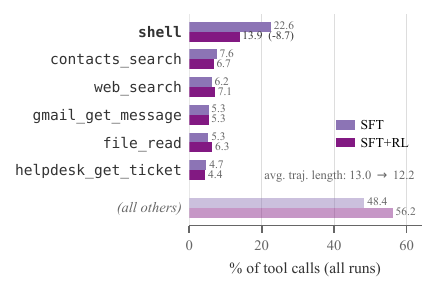}
\caption{ZeroClaw: tool usage}
\label{fig:behavior-tools}
\end{subfigure}
\hfill
\begin{subfigure}[b]{0.38\textwidth}
\centering
\includegraphics[width=\linewidth]{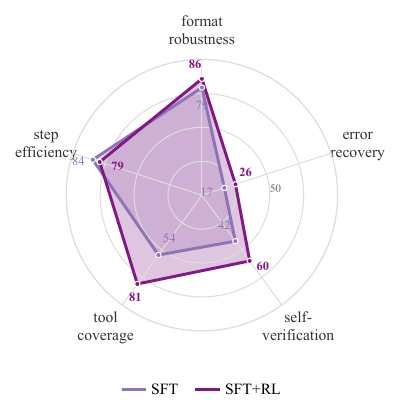}
\caption{Codex: behavioral profile}
\label{fig:behavior-codex}
\end{subfigure}
\caption{SFT vs.\ SFT+RL behavior on ClawEval. RL shifts calls from the generic \texttt{shell} tool to dedicated service tools (a), and improves agentic capbilities such as self-verification (b).}
\label{fig:behavior}
\end{figure}


Finally, we investigate what RL learns on top of SFT.
We run the SFT and SFT+RL \openforgeclaw{} checkpoints on ClawEval and compare 100 trajectories from each, examining tool-call statistics under the ZeroClaw harness (\Cref{fig:behavior}, left) and higher-level behavioral capabilities under the Codex harness (\Cref{fig:behavior}, right).

On ZeroClaw, we find RL changes \emph{how} the model uses tools.
It reduces generic \texttt{shell} calls from 22.6\% to 13.9\% of all tool calls and redistributes them toward dedicated service tools, while also slightly shortening trajectories (\Cref{fig:zeroclaw-tools-full} gives the full breakdown).
This suggests that RL teaches the model to reach for the right specialized tool instead of falling back on a general-purpose shell.

On Codex, we find RL improves several agentic capabilities (see \Cref{subsec:Behavorial Analysis} for detailed definitions) that matter for long-horizon tool use.
In particular, it strengthens error recovery (correctly solving a task after a failed command) and self-verification (reading back its own writes to confirm them), and, echoing the ZeroClaw analysis, it exercises a wider set of the tools each task requires.
These are the behaviors that keep an agent reliable across many steps.
Error recovery, however, remains the weakest capability even after RL.
We hypothesize that such capabilities are difficult to acquire from RL alone, and may require dedicated data or training methods to strengthen further.


\section{Conclusion}
\label{sec:Conclusion}

We present \openforgerl{}, an open framework for training LLM- and VLM-based agents end-to-end, directly inside the inference harnesses they are deployed with.
\openforgerl{} makes \emph{any harness $\times$ any environment} trainable with standard RL codebases such as \verl{}, so agents can be optimized in their real deployment settings rather than in simplified reimplementations.
Using only hundreds to a few thousand automatically curated tasks, we train \openforgeclaw{} and \openforgegui{} models that surpass open models of similar size on nearly all of our tool-use and GUI benchmarks, and in the GUI setting match or exceed models several times larger.
Concretely, \openforgeclaw{} reaches $31.7$~($\mathrm{pass}^3$) on ClawEval, $33.7$ on QwenClawBench, and $28.1$ on MCPAtlas, while \openforgegui{} reaches $37.7$ on OSWorld-Verified, $63.0$ on Online-Mind2Web, and $72.3$ on WebVoyager.
Beyond training, \openforgerl{} lets us analyze how harness choice and RL shape agent behavior, a study prior open work could not easily conduct.
We find that some harnesses are substantially harder to learn than others, that training gains transfer to harnesses unseen during training, and that RL primarily improves \emph{agentic reliability}: the model verifies its own actions, covers more of the tools each task needs, and completes multi-step plans.
Error recovery, however, remains weak even after RL, suggesting that some capabilities may need dedicated data.
We release our code, data, and models, and hope \openforgerl{} lowers the barrier to training and studying agents in the real harnesses and environments where they are deployed.


\bibliography{iclr2026_conference}
\bibliographystyle{iclr2026_conference}

\clearpage
\appendix

\setcounter{table}{0}
\renewcommand{\thetable}{A\arabic{table}}
\setcounter{figure}{0}
\renewcommand{\thefigure}{A\arabic{figure}}

\section{Training Details}
\label{sec:Training Details Appendix}

\subsection{Claw Agent Training Details}
\label{subsec:Claw Training Details}
\begin{figure}[h]
\centering
\includegraphics[width=\textwidth]{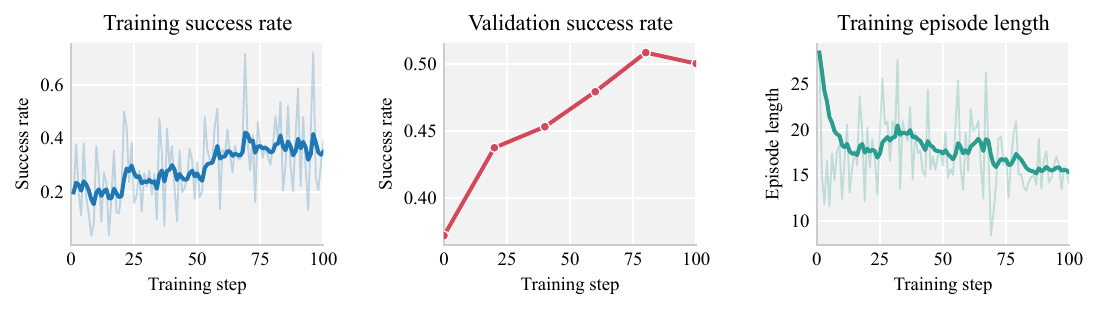}
\caption{\openforgeclaw{} RL training curves on claw tasks. \emph{Left}: training success rate. \emph{Middle}: validation success rate. \emph{Right}: training episode length.}
\label{fig:claw-curve}
\end{figure}

For \openforgeclaw{} RL training, we use \verl{} as the training backend and Microsoft Azure as the cloud provider for rollout containers.
Each rollout runs in its own container, built from a task-specific Dockerfile with the target harness (e.g., OpenClaw, Codex, or ZeroClaw) pre-installed, and is scheduled onto a Kubernetes pod capped at 2 CPUs and 2GB RAM.
These pods are packed onto Azure D128ads~v5 nodes, while policy training runs on a single node of 8$\times$B200 GPUs.
We report the key training hyperparameters in \Cref{tab:training-hparams} and the training curves in \Cref{fig:claw-curve}.

\subsection{GUI Agent Training Details}
\label{subsec:GUI Training Details}
For \openforgegui{} RL training, we similarly use \verl{} as the training backend and Microsoft Azure as the cloud provider for rollout containers.

\begin{figure}[h]
\centering
\includegraphics[width=\textwidth]{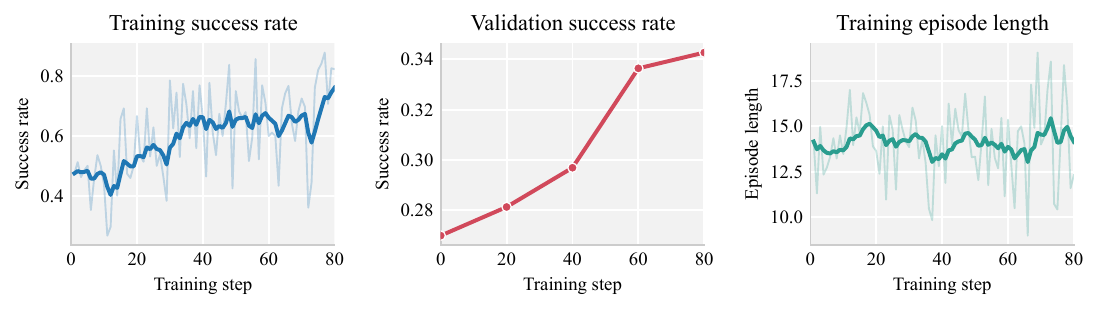}
\caption{\openforgegui{} RL training curves on computer-use tasks. \emph{Left}: training success rate. \emph{Middle}: validation success rate. \emph{Right}: training episode length.}
\label{fig:computeruse-curve}
\end{figure}

\paragraph{Computer-Use} For computer-use agent training, each rollout runs in its own container built from a task-specific Dockerfile with the target harness (a modified version of Kimi-Agent, see \Cref{subsec:computeruse evaluation details}) pre-installed, and is scheduled onto a Kubernetes pod capped at 4 CPUs and 4GB RAM.
These pods are packed onto Azure D64ads~v5 nodes, while policy training runs on a single node of 8$\times$B200 GPUs.
We report the key training hyperparameters in \Cref{tab:training-hparams} and the training curves in \Cref{fig:computeruse-curve}.

\paragraph{Browser-Use} For browser-use agent training, each rollout drives a dedicated remote browser session on the Browser-Use Cloud service (hosted Chromium) over the Chrome DevTools Protocol.
The per-rollout environment, which handles session management, action execution, screenshot capture, and the reward judge, runs in a lightweight Kubernetes sandbox pod capped at 2 CPUs and 6\,GB RAM; the browser itself runs remotely rather than inside the pod.
These pods are packed onto Azure D128ads~v7 nodes, and up to 48 rollouts run concurrently, while policy training runs on a single node of 8$\times$B200 GPUs.
We report the key training hyperparameters in \Cref{tab:training-hparams}.

\begin{table}[ht]
\centering
\caption{Key RL training hyperparameters for \openforgeclaw{} and \openforgegui{}.}
\label{tab:training-hparams}
\small
\setlength{\tabcolsep}{10pt}
\renewcommand{\arraystretch}{1.2}
\begin{tabular}{@{}lccc@{}}
\toprule
\textbf{Hyperparameter} & \textbf{Claw} & \textbf{Computer-Use} & \textbf{Browser-Use} \\
\midrule
Learning rate            & 1e-6  & 5e-7  & 1e-6  \\
Batch size               & $8$ & $8$ & $12$  \\
Group size               & $8$ & $8$ & $5$  \\
KL coefficient           & 0.001  & 0.01  & 0.0  \\
Entropy coefficient      & 0.0  & 0.0  & 0.001  \\
Max rollout time / step  & 900s  & 600s  & 900s  \\
Training steps           & 100  & 80  & 100  \\
Rollout VM CPU / pod     & 2  & 4  & 2  \\
Rollout VM Memory / pod  & 2GiB  & 4GiB  & 6GiB  \\
Total Training Time      & 48H  & 36H  & 32H  \\
\bottomrule
\end{tabular}
\end{table}

\section{Data Details}
\label{sec:Data Details}

\subsection{More Details on Data Synthesis}
\label{subsec:Data Synthesis Pipeline}

We implement our data synthesis pipeline (\Cref{subsec:openforgeenv}) on the Claude Agent SDK, using Opus 4.6 \citep{claude46} as the backbone for each of the five agent modules.
Given a target domain (as a natural-language prompt) and a target number of tasks $N$, the pipeline spawns many agents in parallel that run the following five stages.

\begin{itemize}
    \item \textbf{Propose.} Each agent drafts candidate instructions by browsing the assets and reference pool we construct for the target domain, as we find that directly prompting a model to invent tasks from scratch tends to yield infeasible or unoriginal instructions.
    For Claw agents, this pool comprises (1) skills from ClawHub and (2) tasks from ZClawBench \citep{zclawbench}; for computer-use agents, it comprises (1) X search API access, (2) 22k instructions from AgentNet, and (3) synthetic files and data from Synthetic-Computer-Use.
    To record drafted instructions and avoid duplicates, we maintain a shared SQLite database and instruct each agent to append its proposals to it.
    \item \textbf{Prune.} From the proposed but unimplemented instructions in the database, these agents select the $N$ highest-quality and most diverse instructions, and mark the remainder as discarded.
    \item \textbf{Build.} For each selected instruction, an agent constructs a fully executable environment: tool servers, mock websites, mock data and files, a Dockerfile that packages everything into an image, and a verifier script that scores task success.
    \item \textbf{Test.} Many environment errors and instruction ambiguities are hard to detect without actually running the task. We therefore provide the agent with a script that invokes a separate open LLM/VLM to attempt the task in the built environment: MiniMax-M2.5 for Claw tasks and Kimi-K2.5 for computer-use tasks.
    \item \textbf{Refine.} Finally, the refine agent inspects the rollout trajectory and the verifier's score and decides whether the environment has any defects or the instruction is ambiguous. If so, it patches the environment and/or revises the instruction, and repeats the test-and-refine loop until the task passes all checks.
\end{itemize}

\subsection{More Details on Claw Data}
\label{subsec:More Details on Claw Data}

\begin{figure}[t]
\centering
\includegraphics[width=0.9\textwidth]{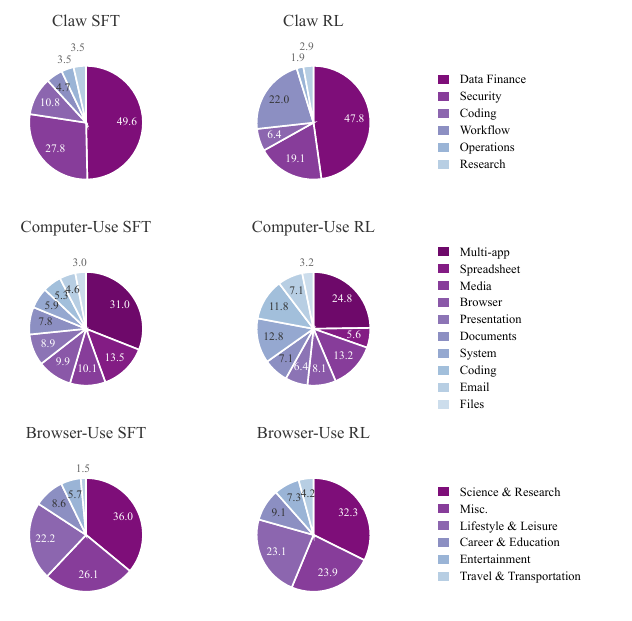}
\caption{Category distribution (\%) of SFT and RL training tasks used in the Claw (top), Computer-Use (middle), and Browser-Use (bottom) domains.}
\label{fig:dset-distribution-full}
\end{figure}

At the time of the project, we did not find any large scale public dataset suitbale for training Claw-based agents, especially for RL training.
As a result, we primarily rely on our data synthesis pipeline (\Cref{subsec:Data Synthesis Pipeline}) to generate tasks for SFT and RL training.
While powerful, we find synthesizing a task with a thoroughly tested verifier is costly in both time and money, as it involves multiple rounds of real rollouts, refinement, and re-rollout.
To save cost, we instead consider for \emph{SFT tasks}, we skip the test and refinement stage, and directly prompt GPT-5.4 as a judge to determine task success. 
This is much afforadable and reasonable since for SFT trajectories the success signals are often only used for onetime data filtering.
For RL tasks, we maintain the full test-and-refine loop to ensure the verifier is robust and reliable, as RL training requires repeated rollouts and the verifier must be able to consistently judge task success.
On average, synthesizing an RL task (with verifier and refining) takes 16.1 minutes and 4.36 USD, while synthesizing an SFT task (without verifier) takes 5.2 minutes and 0.86 USD.
A full breakdown of our task distribution is shown in \Cref{fig:dset-distribution-full}.

For harnesses, in addition to the default \react{} loop implemented by most tool-use agents, we also include three popular harnesses: ZeroClaw, OpenClaw, and Codex, to experiment with the capabilities of our \openforgerl{} infrastructure.

\subsection{More Details on GUI Data}
\label{subsec:More Details on GUI Data}

\paragraph{Computer-use tasks} At the time of this work, there was no large-scale public dataset for training computer-use agents, especially with RL.
One reason is that running a computer-use GUI in a lightweight container is challenging: such environments are commonly run as full Ubuntu virtual machines under QEMU \citep{xie2024osworldbenchmarkingmultimodalagents}.
We find, however, that Xvfb can instead render a virtual display in memory, which is far more lightweight and lets us run many containers in parallel for RL training.
With computer-use GUIs now runnable in lightweight containers, we follow our Claw setup and use the same data synthesis pipeline (\Cref{subsec:Data Synthesis Pipeline}) to generate SFT and RL tasks.
For \emph{SFT tasks}, we again skip the test-and-refine stages and prompt GPT-5.4 as a judge to determine task success.
On average, each SFT task takes 4.0 minutes and 1.37 USD to synthesize.
For \emph{RL tasks}, we keep the full test-and-refine loop so that the verifier is robust and reliable.
On average, synthesizing an RL task (with verification and refinement) takes 21.3 minutes and 6.12 USD.

As the harness, we primarily use Kimi-Agent \citep{kimiteam2026kimik25visualagentic}, following the implementation of \citet{xie2024osworldbenchmarkingmultimodalagents}, with slight modifications to support additional tools such as bash, following Anthropic's computer-use approach. For the full action space and tool list, see \Cref{subsec:computeruse evaluation details}.

\paragraph{Browser-use tasks}
For browser-use, we adapt MolmoWeb~\citep{gupta2026molmowebopenvisualweb}'s codebase as the harness and our main modifications involve (1) using json-formatted action space other than MolmoWeb's tool-call format to avoid extra training stages for action alignment, and (2) integrating Browser-Use Stealth Browsers \citep{browseruse} following OpenWebRL to solve CAPTCHA and website blockings, which we found reducing the ratio of IP and CAPTCHA block from 40\% to nearly zero. We provide a more detailed description and examples of the observation and action spaces for both environments in \Cref{subsec:browseruse evaluation details}.

For SFT data, we follow a similar data curation pipeline to OpenWebRL including task-filtering and trajectory collection. We start from subsampling the PAE-WebVoyager~\citep{PAE} split from WebGym~\citep{bai2026webgymscalingtrainingenvironments} by removing tasks that overlap with evaluation benchmarks and subtasks decomposed from parent intents tasks. Then, we keep only the most popular websites that exists in SimilarWeb Top100 and MOZ Top 500 to remove the distrations of unusual websites in the long-tail distribution. Finally, to remain the diversity of tasks, we embed task instructions with
Qwen3-Embedding-8B \citep{zhang2025qwen3embeddingadvancingtext} and apply greedy similarity-based deduplication with a predefined threshold 0.55. The eventual SFT candidate pool contains 2500 tasks.

We apply a stronger teacher model (Kimi-K2.5) to infer on the candidate pool with max turns 30 and 4 repeats. We collect all successful trajectories, and keep only the shortest trajectory among multiple success of a single task. We found that repeated actions commonly appear for complex websites, and we reserve only the last turn for more than 3 consecutive identical actions anf remove the entire trajectory if more than 5 identical actions occur in the process. Following the above operations, we curate 1496 trajectories for distillation.  

During RL stage, we apply the same popular website filter as SFT and randomly sample 600 tasks from the Insta-v3 \citep{Trabucco2025InSTA} and 300 tasks from the PAE-WebVoyager \citep{PAE} split that are distinct from the SFT pool. The maximum turn is restricted to 20 to balance the performance and training cost.

\section{Evaluation Details}
\label{sec:Evaluation Details Appendix}

\subsection{ClawEval and QwenClawBench Evaluation Details}
\label{app:claweval details appendix}
%
We evaluate our \openforgeclaw{} models on popular claw- and harness-related benchmarks that measure how well an agent can use a harness to solve tasks.
Specifically, we use ClawEval \citep{ye2026clawevaltrustworthyevaluationautonomous}, QwenClawBench \citep{qwenclawbench1.1}, and MCPAtlas \citep{bandi2026mcpatlas}, the last serving as a related but ``held-out'' test set for novel tool use.
All results in our main experiments (\Cref{subsec:Claw Agents}) use each benchmark's official evaluation protocol: for ClawEval, the official \react{}-like loop repeatedly prompts the model to reason and call tools; for QwenClawBench, the OpenClaw harness is used.

To study the effect of harness choice on model performance, we additionally evaluate on ClawEval under three other harnesses: ZeroClaw, OpenClaw, and Codex (\Cref{subsec:Cross Harness Comparison}).
Because ClawEval requires the model to call custom tool servers, ZeroClaw was straightforward to support, as it natively allows registering new tools.
To adapt OpenClaw and Codex, which is non-trivial for adding custom tools, we expose each ClawEval-specific tool to them through \texttt{SKILL.md} files.
Specifically, for each custom tool server, we prompt an LLM (Claude Opus 4.6) with its API signature and tool descriptions to generate a \texttt{SKILL.md} file that explains how to call the tool from a bash command line.
At inference time, these files are loaded into the harness, and the model invokes the tools through the bash interface.
Evaluation protocols are unchanged from the official ClawEval benchmarks.

\subsection{MCP-Atlas Evaluation Details}
\label{app:mcpatlas}

For reproducibility, we evaluate MCPAtlas \citep{bandi2026mcpatlas} under the benchmark's default 20-server configuration, which excludes optional servers that require third-party credentials or service-specific data initialization.
This configuration fixes the evaluation set without any manual selection on our part: of the 500 public tasks, exactly 89 have ground-truth expected tool calls that are fully supported by these default servers.
We evaluate every model on this same 89-task set, holding the task identifiers and environment configuration fixed.

To evaluate, we use the MCPAtlas official harness which is based on a \react{}-like loop: the harness exposes the task-specific MCP tools to the policy model, executes its tool calls in the MCPAtlas sandbox, and returns the resulting observations until the model terminates.
We use the default task prompts and tool configurations, and add no benchmark-specific demonstrations or fine-tuning.

Following the MCPAtlas claim-coverage protocol, we score each final response against the ground-truth factual claims using Gemini 2.5 Pro as the judge.
A task counts as successful when its claim-coverage score is at least $0.75$, and we report the fraction of successful tasks as pass@1.
Every model is evaluated under the same task subset, harness, judge, and threshold.

\subsection{Computer-Use Evaluation Details}
\label{subsec:computeruse evaluation details}
%
For computer-use environments, we evaluate on OSWorld-Verified \citep{xie2024osworldbenchmarkingmultimodalagents}.
OSWorld is a popular benchmark that evaluates how well a multimodal agent can complete open-ended, real-world computer tasks on a real Ubuntu desktop, operating applications through the screen with mouse and keyboard and being scored by task-specific, execution-based verifiers.
OSWorld-Verified is an in-place upgrade of OSWorld with enhanced infrastructure and improved task quality built from 300+ pieces of feedback from the community.

\begin{figure}[t]
\centering
\setlength{\fboxsep}{8pt}
\fbox{\begin{minipage}{\dimexpr\textwidth-2\fboxsep-2\fboxrule\relax}
\small

\textbf{Task instruction.}\ Go to the second slide and name its title as ``Online Shopping'' with the same color, position, and font size as the previous title.

\begin{center}
\includegraphics[width=0.72\linewidth]{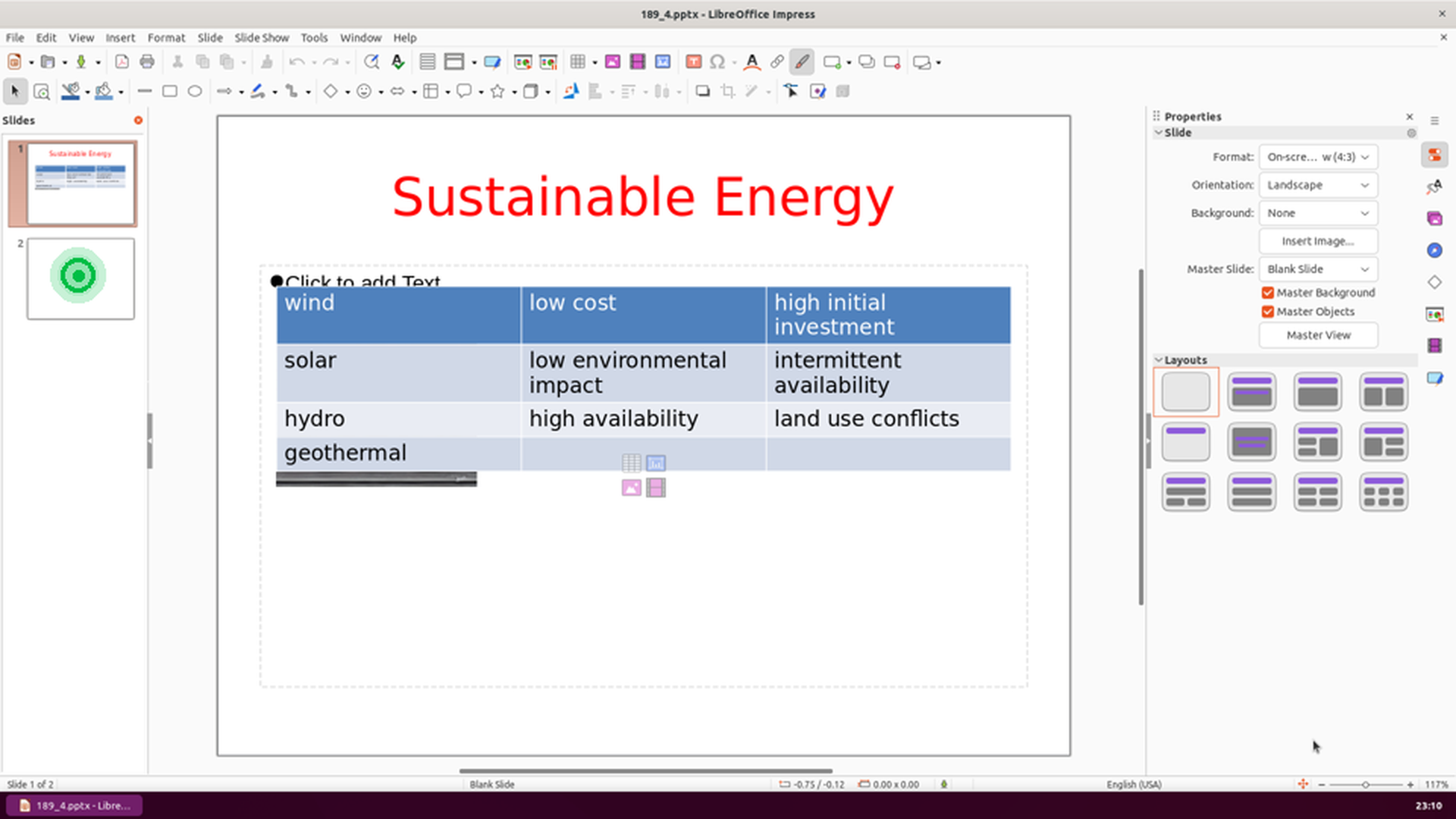}
\end{center}

\textbf{Model output.}\\
\textless think\textgreater\\
The user wants me to go to the second slide and set its title to ``Online Shopping'' with the same formatting as the current title ``Sustainable Energy''. The current title is red, in a large font, at the top center. On the left, a slide panel shows ``Slide 1 of 2'', and slide 2 appears blank. I should click on slide 2 to navigate to it first, then add the title with the same formatting. $[\dots]$\\
\textless/think\textgreater\\
\textless tool\_call\textgreater\\
\{``name'': ``computer'', ``arguments'': \{``action'': ``left\_click'', ``coordinate'': [0.054, 0.337]\}\}\\
\textless/tool\_call\textgreater
\end{minipage}}

\caption{An example computer-use input-output pair. Given the task instruction and the current screenshot, the model emits a single generation with reasoning (\texttt{<think>}, truncated here) and a tool call (\texttt{<tool\_call>}). To improve readability, in this example we overlay the predicted normalized coordinate with a \textcolor{teal}{green marker} on the screenshot.}
\label{fig:computeruse-example}
\end{figure}

Evaluating on OSWorld-Verified first requires choosing a harness.
Because our computer-use data pipeline is built around Kimi-K2.5 \citep{kimiteam2026kimik25visualagentic}, we use Kimi-Agent, as implemented in the official OSWorld repository, as our main training and evaluation harness.
To make the harness more efficient, we add a lightweight modification inspired by Anthropic's computer-use approach: we additionally expose \texttt{bash} and a \texttt{str\_replace\_editor} tool, letting the model directly manipulate files and data (e.g., editing text, CSV, and JSON files) without opening a GUI application.
We run Kimi-Agent in screenshot-only mode: every observation is a screenshot of the GUI, with no additional text or metadata.
We show an example trajectory in \Cref{fig:computeruse-example} and the full action space and tool list in \Cref{tab:computeruse-actions}.

\begin{table}[t]
\centering
\caption{Action space of our computer-use agent. The agent controls an Ubuntu desktop through three tools: \texttt{computer} (mouse, keyboard, scrolling, and episode control), \texttt{bash} (shell access), and \texttt{str\_replace\_editor} (file editing). Coordinates are normalized to $[0,1]$ relative to the screenshot ($x{=}0$ left, $x{=}1$ right, $y{=}0$ top, $y{=}1$ bottom), backed by a $1920\times1080$ screen.}
\label{tab:computeruse-actions}
\small
\setlength{\tabcolsep}{4pt}
\renewcommand{\arraystretch}{1.15}
\begin{tabular}{@{}p{0.30\textwidth}p{0.21\textwidth}p{0.41\textwidth}@{}}
\toprule
\textbf{Action / command} & \textbf{Arguments} & \textbf{Description} \\
\midrule
\multicolumn{3}{@{}l}{\textit{\quad\texttt{computer} --- desktop mouse, keyboard, scrolling, and control}}\\
\cmidrule(lr){1-2}
\addlinespace[2pt]
\raggedright \texttt{left\_click}, \texttt{right\_click}, \texttt{middle\_click}, \texttt{double\_click}, \texttt{triple\_click}, \texttt{left\_press}
  & \raggedright \texttt{coordinate}, \texttt{keys}
  & \raggedright Click at \texttt{coordinate} with the given button and click count; \texttt{left\_press} presses and holds briefly. \texttt{keys} adds modifiers. \tabularnewline
\raggedright \texttt{mouse\_move}, \texttt{left\_click\_drag}, \texttt{left\_mouse\_down}, \texttt{left\_mouse\_up}
  & \raggedright \texttt{coordinate}, \texttt{start\_coordinate}, \texttt{coordinate2}
  & \raggedright Move the cursor, drag from \texttt{start\_coordinate} to \texttt{coordinate2}, or press/release the left button for manual drags. \tabularnewline
\raggedright \texttt{key}, \texttt{type}, \texttt{hold\_key}, \texttt{key\_down}, \texttt{key\_up}
  & \raggedright \texttt{text}, \texttt{keys}, \texttt{duration}
  & \raggedright Type \texttt{text}, press a key or hotkey (e.g., \texttt{ctrl+l}), or hold / press / release specific \texttt{keys}. \tabularnewline
\raggedright \texttt{scroll}, \texttt{hscroll}
  & \raggedright \texttt{scroll\_direction}, \texttt{scroll\_amount}, \texttt{coordinate}
  & \raggedright Vertical or horizontal mouse-wheel scroll by \texttt{scroll\_amount} in \texttt{scroll\_direction}, optionally anchored at \texttt{coordinate}. \tabularnewline
\raggedright \texttt{screenshot}, \texttt{wait}, \texttt{terminate}, \texttt{done}, \texttt{fail}
  & \raggedright \texttt{duration}, \texttt{status}
  & \raggedright Capture the screen, wait, or end the episode as success / failure (\texttt{status}). \tabularnewline
\midrule
\multicolumn{3}{@{}l}{\textit{\quad\texttt{bash} --- shell access}}\\
\cmidrule(lr){1-2}
\addlinespace[2pt]
\raggedright \texttt{bash}
  & \raggedright \texttt{command}, \texttt{timeout}, \texttt{working\_dir}
  & \raggedright Run a shell \texttt{command} inside the desktop environment (not the host). \tabularnewline
\midrule
\multicolumn{3}{@{}l}{\textit{\quad\texttt{str\_replace\_editor} --- file viewing and editing}}\\
\cmidrule(lr){1-2}
\addlinespace[2pt]
\raggedright \texttt{view}, \texttt{create}, \texttt{str\_replace}, \texttt{insert}, \texttt{undo\_edit}
  & \raggedright \texttt{path}, \texttt{file\_text}, \texttt{old\_str}, \texttt{new\_str}, \texttt{insert\_line}, \texttt{view\_range}
  & \raggedright View, create, and edit files with precise text operations at absolute \texttt{path}s. \tabularnewline
\bottomrule
\end{tabular}
\end{table}


\subsection{Browser-Use Evaluation Details}
\label{subsec:browseruse evaluation details}
\begin{figure}[t]
\centering
\setlength{\fboxsep}{8pt}
\fbox{\begin{minipage}{\dimexpr\textwidth-2\fboxsep-2\fboxrule\relax}
\small

\textbf{Task instruction.}\ Find a pair of wireless headphones on Amazon with active noise canceling for \$100 or less and add them to the cart.

\begin{center}
\includegraphics[width=0.78\linewidth]{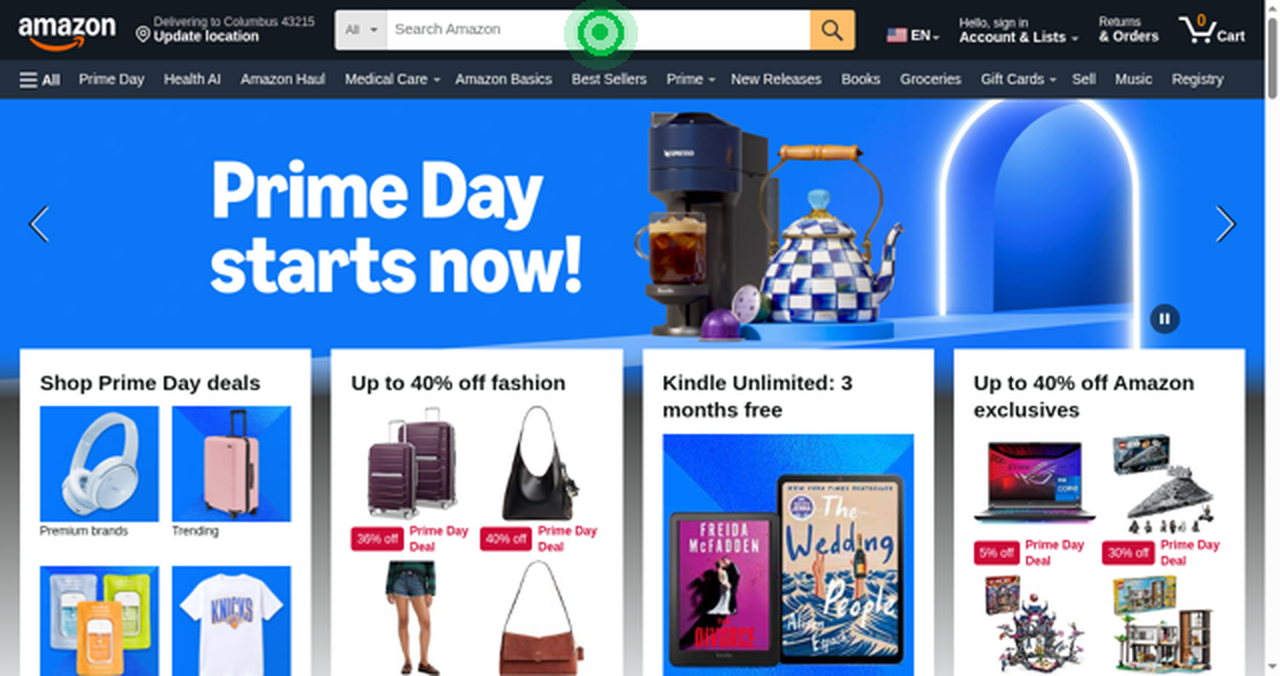}
\end{center}

\textbf{Model output.}\\
\textless think\textgreater\\
Good, I'm now on the Amazon homepage. I need to search for wireless headphones with active noise canceling for \$100 or less. I'll click on the search bar and type my search query.\\
\textless/think\textgreater\\
\textless tool\_call\textgreater\\
\{``name'': ``computer\_use'', ``arguments'': \{``action'': ``left\_click'', ``coordinate'': [0.469, 0.048]\}\}\\
\textless/tool\_call\textgreater
\end{minipage}}

\caption{A representative browser-use step on an Online-Mind2Web task. Given the task instruction and the current screenshot, the model emits a single generation that interleaves its reasoning (\texttt{<think>}) and a tool call (\texttt{<tool\_call>}). We overlay the predicted normalized click coordinate with a \textcolor{teal}{green marker} on the screenshot, which correctly targets the Amazon search bar.}
\label{fig:browser-example}
\end{figure}

For browser-use, we evaluate on Online-Mind2Web \citep{onlinemine2web} (300 tasks) and WebVoyager \citep{he2024webvoyagerbuildingendtoendweb}, the latter on the 595-task ``Fara-595'' subset released with Fara-7B \citep{fara7b2025}.
We use our modified MolmoWeb harness (\Cref{subsec:More Details on GUI Data}), which exposes a single \texttt{computer\_use} tool; the full action space is listed in \Cref{tab:browser-actions}.
Observations are screenshot-only: each step provides the current $1280\times720$ viewport screenshot, with no DOM or accessibility tree, together with a short text block giving the previous action's result, the page title and URL, and the step count.
Click and drag coordinates are normalized to $[0,1]$ relative to the screenshot, the same convention as our computer-use agent.
At inference, we cap each episode at 30 steps, with a 900\,s per-task and 90\,s per-action timeout, and score Online-Mind2Web with the AgentTrek protocol (o4-mini) and WebVoyager with the official protocol (GPT-4o).
We show an example step in \Cref{fig:browser-example}.

\begin{table}[t]
\centering
\caption{Action space of our browser-use agent. The model controls a Chromium browser through a single \texttt{computer\_use} tool. Click and drag coordinates are normalized to $[0,1]$ relative to the screenshot, which is pinned to a $1280\times720$ viewport.}
\label{tab:browser-actions}
\small
\setlength{\tabcolsep}{4pt}
\renewcommand{\arraystretch}{1.15}
\begin{tabular}{@{}p{0.30\textwidth}p{0.21\textwidth}p{0.41\textwidth}@{}}
\toprule
\textbf{Action} & \textbf{Arguments} & \textbf{Description} \\
\midrule
\multicolumn{3}{@{}l}{\textit{\quad\texttt{computer\_use} --- browser mouse, keyboard, scrolling, navigation, and control}}\\
\cmidrule(lr){1-2}
\addlinespace[2pt]
\raggedright \texttt{left\_click}, \texttt{right\_click}, \texttt{middle\_click}, \texttt{double\_click}, \texttt{triple\_click}
  & \raggedright \texttt{coordinate}
  & \raggedright Click at \texttt{coordinate} with the given button and click count (double/triple select a word/line). \tabularnewline
\raggedright \texttt{mouse\_move}, \texttt{left\_click\_drag}
  & \raggedright \texttt{coordinate}
  & \raggedright Move the cursor to \texttt{coordinate}, or drag from the current cursor position to it. \tabularnewline
\raggedright \texttt{type}, \texttt{key}
  & \raggedright \texttt{text}, \texttt{keys}
  & \raggedright Type \texttt{text} into the focused field, or press a key or chord (e.g., \texttt{["ctrl","a"]}). \tabularnewline
\raggedright \texttt{scroll}, \texttt{hscroll}
  & \raggedright \texttt{pixels}
  & \raggedright Vertical or horizontal scroll at the cursor by \texttt{pixels} (sign gives direction). \tabularnewline
\raggedright \texttt{goto}, \texttt{go\_back}, \texttt{new\_tab}, \texttt{tab\_focus}
  & \raggedright \texttt{url}, \texttt{index}
  & \raggedright Navigate to \texttt{url}, go back in history, open a new tab, or switch to the tab at \texttt{index}. \tabularnewline
\raggedright \texttt{wait}, \texttt{terminate}, \texttt{answer}
  & \raggedright \texttt{time}, \texttt{status}, \texttt{text}
  & \raggedright Wait for \texttt{time} seconds, end the episode (\texttt{terminate} with success / failure), or return the final \texttt{answer}. \tabularnewline
\bottomrule
\end{tabular}
\end{table}

\begin{figure}[t]
\centering
\includegraphics[width=\textwidth]{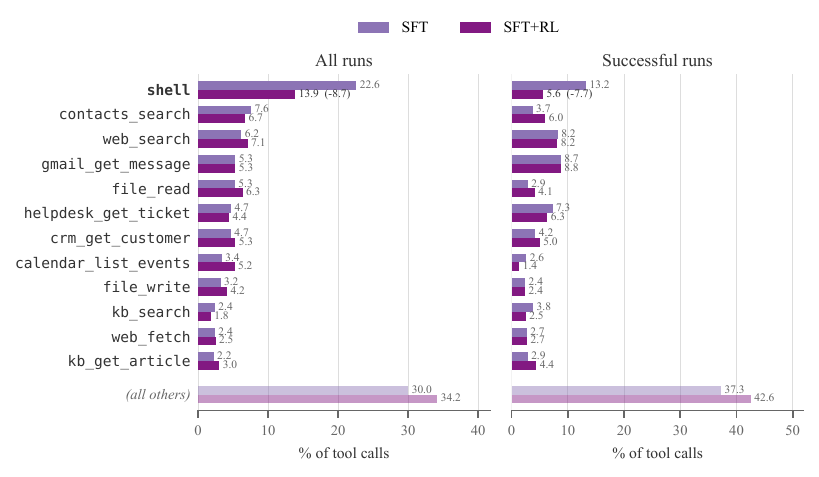}
\caption{Full tool-usage distribution on ClawEval with the ZeroClaw harness for SFT vs.\ SFT+RL, over all runs (left) and successful runs (right). Complements \Cref{fig:behavior}: the shift away from the generic \texttt{shell} tool is even stronger when restricted to successful runs (13.2\% $\rightarrow$ 5.6\%).}
\label{fig:zeroclaw-tools-full}
\end{figure}

\section{Analysis Details}
\label{sec:Analysis Details}

\subsection{Behavioral Analysis Details}
\label{subsec:Behavorial Analysis}

We analyze behavioral differences between the SFT and SFT+RL \openforgeclaw{} checkpoints on ClawEval, running each checkpoint under the ZeroClaw and Codex harnesses.

\paragraph{Tool usage (ZeroClaw).}
For each tool, we report the percentage of all tool calls that invoke it, aggregated over all rollouts.
This captures which tools the model relies on, independent of whether a task is ultimately solved.

\paragraph{Behavioral capabilities (Codex).}
We summarize five capabilities relevant to long-horizon tool use, shown as the radar plot in \Cref{fig:behavior} (right).
Each is a percentage in $[0, 100]$ where higher is better, computed over rollouts from the two \openforgeclaw{} checkpoints on the same set of tasks.

\begin{itemize}
    \item \textbf{Format robustness.} The percentage of rollouts \emph{not} terminated by a malformed tool call, i.e., $100$ minus the fraction of rollouts in which the harness fails to parse the model's function-call payload (typically a long structured output) and the session disconnects. It measures how reliably the model emits well-formed tool calls.
    \item \textbf{Error recovery.} Among rollouts that hit at least one failed command, the percentage that still solve the task. It measures whether the model can proceed after an error rather than crashing or giving up.
    \item \textbf{Self-verification.} The percentage of write (state-changing) tool calls that are followed by a read-back of the same service, e.g., listing tasks after creating one. It measures whether the model checks the effect of its own actions.
    \item \textbf{Tool coverage.} On tasks that require at least three distinct services, the percentage of rollouts that invoke \emph{every} required service at least once. It measures whether the model carries a multi-part plan through to completion.
    \item \textbf{Step efficiency.} For each task, we define the \emph{best} step count as the fewest steps among all successful runs (from 3 runs per task) of that task across both checkpoints. A checkpoint's score on a task it solves is $\text{best} / \text{(its own step count)}$, averaged over its solved tasks. A higher value means the model solves tasks in closer to the minimum number of steps.
\end{itemize}

\end{document}